\documentclass[sigchi]{acmart}
\renewcommand\footnotetextcopyrightpermission[1]{} 
\AtBeginDocument{%
  \providecommand\BibTeX{{%
    \normalfont B\kern-0.5em{\scshape i\kern-0.25em b}\kern-0.8em\TeX}}}

\usepackage{multirow}
\usepackage{graphicx}
\usepackage{subcaption}
\graphicspath{ {./figures/} }
\usepackage{hyperref}

\begin{document}

\title{Detecting Handwritten Mathematical Terms with Sensor Based Data}

\author{Lukas Wegmeth}
\affiliation{%
  \institution{Ubiquitous Computing,\\ University of Siegen}
  \streetaddress{H\"olderlinstr. 3}
  \postcode{57076}
}
\email{lukas.wegmeth@student.uni-siegen.de}
\author{Alexander Hoelzemann}
\affiliation{%
  \institution{Ubiquitous Computing,\\ University of Siegen}
  \streetaddress{H\"olderlinstr. 3}
  \postcode{57076}
}
\email{alexander.hoelzemann@uni-siegen.de}
\author{Kristof Van Laerhoven}
\orcid{0000-0001-5296-5347}
\affiliation{%
  \institution{Ubiquitous Computing,\\ University of Siegen}
  \streetaddress{H\"olderlinstr. 3}
  \postcode{57076}
}
\email{kvl@eti.uni-siegen.de}

\begin{abstract}
In this work we propose a solution to the UbiComp 2021 Challenge by Stabilo in which handwritten mathematical terms are supposed to be automatically classified based on time series sensor data captured on the DigiPen. The input data set contains data of different writers, with label strings constructed from a total of 15 different possible characters. The label should first be split into separate characters to classify them one by one. This issue is solved by applying a data-dependant and rule-based information extraction algorithm to the labeled data. Using the resulting data, two classifiers are constructed. The first is a binary classifier that is able to predict, for unknown data, if a sample is part of a writing activity, and consists of a Deep Neural Network feature extractor in concatenation with a Random Forest that is trained to classify the extracted features at an F1 score of >90\%. The second classifier is a Deep Neural Network that combines convolution layers with recurrent layers to predict windows with a single label, out of the 15 possible classes, at an F1 score of >60\%. A simulation of the challenge evaluation procedure reports a Levensthein Distance of 8 and shows that the chosen approach still lacks in overall accuracy and real-time applicability.
\end{abstract}

\renewcommand{\shortauthors}{Wegmeth, Hoelzemann and Van Laerhoven}
\maketitle
\pagestyle{plain}
\section{Introduction}
The classification of sensor data through Deep Learning approaches is a well-researched topic within the scientific community, alas still being in its early phases. While some standardised workflows exist and a state-of-the-art has been accomplished and still evolves, there usually remain parts for each specific task where specially handcrafted solutions are required on top of community knowledge.\\
This work emerged from Stabilo's Ubicomp 2021 Challenge revolving around sensor data captured with the DigiPen, a pen equipped with multiple sensors but still reminiscent of a regular ball-point pen. The intent of the challenge is to find a suitable way to classify handwritten terms of writers with unknown writing styles. Naturally, Deep Learning appears to be a suitable approach to a classification problem like this.\\
To accomplish the task, a larger data set of different writers and according labels is given. The problem itself can generally be divided into two larger sub-problems with one being the segmentation of the terms into singular symbols and the other being the classification of said symbols and thereafter the possibility to classify whole terms.\\

\section{Related Work}
The problem presents itself as a combination of handwritten character recognition as well as time series data analysis. The methodology for this work required the application of some well researched concepts. As such, a multitude of superficially similar articles were explored, especially in the context of neural networks.\\
Handwritten character recognition, albeit still a challenging research topic, has become fairly popular in the recent advent of powerful neural network architectures, of course mainly due to its practicability. Traditionally, handwritten characters are processed and recognized through (gray scale) images. Starting more than 20 years ago along the work of \citep{lecun1995learning}, neural networks for digits, characters and other symbols started to evolve rapidly. Today, for different sets of characters, digits and other symbols, there are highly specialised networks available \citep{chatterjee2019bengali} along with large open data sets \citep{lecun-mnisthandwrittendigit-2010} as well as meta-research over these topics \citep{5277565}. Although the problem posed in this work does not implicitly offer image data, the concepts employed by the aforementioned types of research can simplify analogous steps and sometimes also be translated seamlessly.\\
Sensor data analysis, as a topic, is possibly one of the widest in the spectrum of pattern recognition. Sensors are comparatively cheap, usually lightweight in terms of their fingerprint, and are ubiquitous today. To solve the problem formulated in this work, a data set containing non-optical sensor data over time with variable intervals is given, which, in turn, moves the problem into the domain of time series data analysis. Similar to handwritten character recognition, this topic, in conjunction with neural networks, received popular scientific attention over the recent years. There are a multitude of practical applications for time series data analysis, e.g. for natural phenomenon \citep{agrawal2012application} \citep{JAIN2007585}, seasonal variations \citep{ZHANG2005501}, industrial processing \citep{hsu2021multiple} and overarching sensor data analysis in general \citep{8437249}. Some amount of practical knowledge gathered from different approaches in entirely different application domains can be translated into useful solutions in our work, largely due to the similarity of time series sensor data as a whole.\\
This focus of this work is the interaction and combination of the two previously introduced fields. Historically, handwritten characters would be recognized by classifying rasterised images of them. In recent years however, new options for classifying human interaction have emerged, enabled through the continuous evolution of microchips. One of these options are commercially available motion sensing devices and it has been shown that these may be used to track handwriting in live environments \citep{6473522}. Another option are writing tools equipped with different types sensors. Such pens may have a multitude of sensors for different tasks and sometimes run on battery and transmit data by radio or with a wired connection, but otherwise, from the outside, closely resemble a pen as it is usually known. One of the probable goals of equipping a pen with sensors may be the automated recognition of what the user is writing. This is the goal of this work. Many similar works can be found, using either a subset of the sensors \citep{6020787} available to us or additional sensory modalities \citep{10.1145/3173574.3173705}. Related work has shown that handwritten digit recognition through sensor based data may generally be possible. This work aims to provide additional scientific insight to these fields by working with a higher number of sensors as well as non-calibrated input data and solve the underlying challenge task in the process.\\
Previously, Stabilo already released a similar challenge for UbiComp 2020. Contrary to the one discussed in this work, only written letters were to be classified. These were also already segmented, so it required comparatively less focus on data cleanup and preprocessing in general. We have also worked on that challenge, but did not release a public paper on it. However, some other research teams released their approaches in various publications, gathering results by using networks similar to ours for this challenge \citep{DBLP:journals/corr/abs-2008-01078} and finding improvements gathered through domain knowledge extensions \citep{9257740}. They report limited success given the hardware and applicability of the data but show that there is a possibility to classify the data given some additional constraints. We have found similar ideas and pursued the integration of topical knowledge gained from related work and own work into this research.

\section{Methodology}
Generally, applying a Supervised Deep Learning approach to classify time series data is a straightforward workflow since popular procedures are very well documented and have been proven to be applicable in many different domains. This work is no exception to that rule. However, we realized that the posed problem also requires specially handcrafted solutions for some steps along the pipeline and we show that, as it is custom for many data analysis problems, the data cleanup and preprocessing require most care and time. This chapter will outline the general workflow and dive in to the more specialised approaches that were taken to solve the problem given the available input. Some details like the ranges of sensors are not explained in this paper, but rather taken as a given and such details as well as other useful information can be directly accessed through the official site of the challenge (\href{https://stabilodigital.com/ubicomp-2021-challenge/}{https://stabilodigital.com/ubicomp-2021-challenge/}).

\subsection*{Input data and objectives}
\begin{figure}[ht]
    \caption{The raw input data shape. Each person is separated and contains a varying amount of data-label pairs. The length of the data and labels is also not fixed.}
    \centering
    \includegraphics[width=\linewidth]{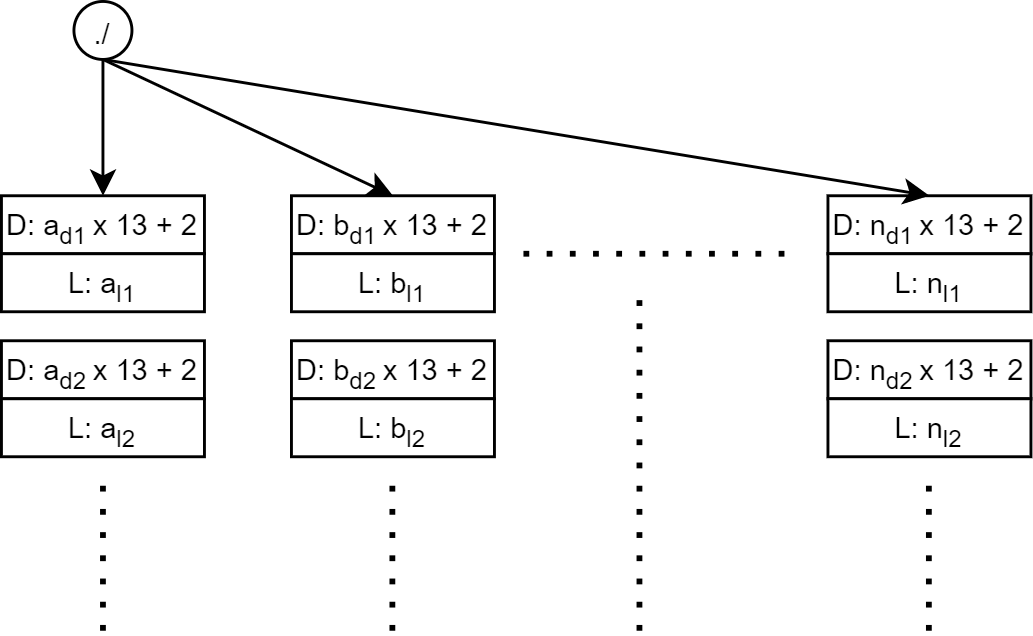}
    \label{input_data}
\end{figure}

The goal is to classify any unknown time series of sensor data. The classes are given by the input characters which include all single-digit Arabic numerals as well as the most common operators, totalling to 15 possible classes. The data was captured with five different sensors, with four of them providing three-dimensional output, resulting in a total of 13 separate sensor tracks per sample as well as two additional tracks for indexing and time interval. The label is a single string built from the given characters with variable length per input data stream. The data is also separated by persons. The general shapes of the input data-label pairs are visualized in figure \ref{input_data}.

Given their shape and the formulated problem, the first issue to solve is assigning each distinct symbol in the label to a region in the time series data to be able to assign a single class to a specific region and classify said region.

\subsection*{Label splitting and preprocessing}
\begin{figure}[ht]
    \caption{The label splitting workflow. Flow diagram for the rule-based label splitting algorithm. Used to accurately segment characters in labelled data.}
    \centering
    \includegraphics[width=\linewidth]{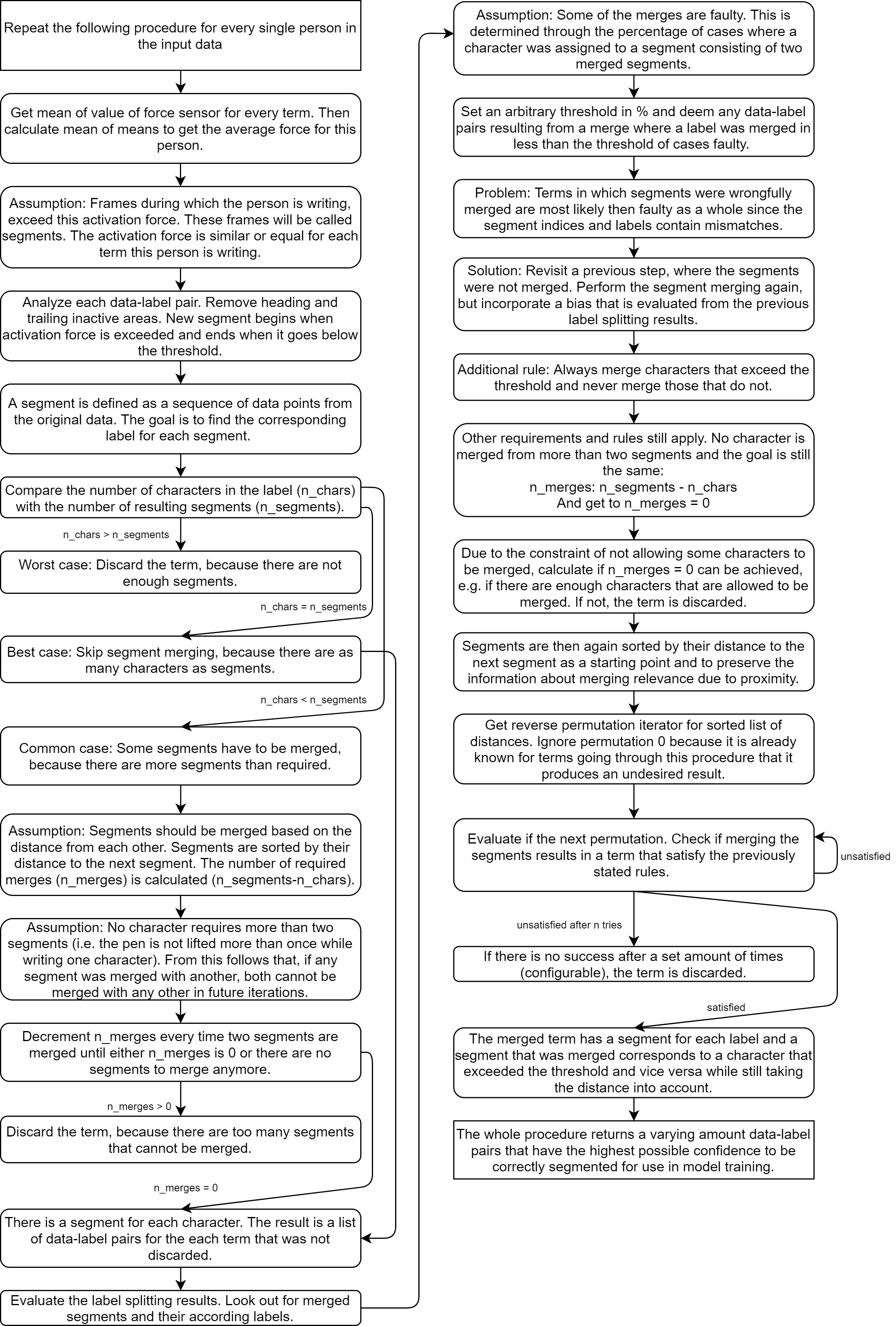}
    \label{label_splitting}
\end{figure}

Without the previously outlined task that we refer to as label splitting, the problem would be ill-defined since there are an exceedingly large amount of possible permutations for 15 possible characters and string lengths of at least 10 and up to 20 characters, where each permutation would be defined as a class in a Deep Learning problem. There is neither enough data to justify this approach nor would it be feasible for any network to learn that many classes even if there were.\\
Since there is no additional information on the location of these regions given, we have to deduct an approach by which they can be separated. We present a rule-based algorithm to perform label splitting for labelled data and a Machine Learning approach to identify them on unlabelled data. The training input for the former stems from a rule-based algorithm that is based on domain knowledge and is applied to the given labelled data.\\
Through experiments we have found that one particular sensor stands out in telling information about the boundary between two separate labels in the sensor data. That is the force sensor, which measures the force applied to the tip of the pen and, by extension, tells us when the tip of the pen was in contact with the paper. This information, however, is not sufficient as simulations have shown that some characters may be written by removing the pen from the paper to draw a new line or curve where at least either the start or end point are disconnected from any point that was written before picking up the pen. The most notable example is the division (:) sign. Writers were instructed to draw this sign as two separated dots which is the German notation. There are other examples like the plus (+) sign or some numerals like 4, 5 or 7, but, other than with division, there is technically no requirement to pick up the pen to write these. Any observation is simply of statistical nature. In order to avoid manually making up inaccurate rules about which characters require the pen to be removed from the paper during the writing process, we would seek an algorithm to accurately report this information for each writer separately to then be able to perform the label splitting task with very high certainty of it being correct.\\
The workflow to achieve this is based on a rule-based algorithm which analyses, modifies and reads the input data in certain ways to detect the writing style of each person separately. The essence of the algorithm is described in the diagram in figure \ref{label_splitting}. Smaller steps are omitted and most of the thresholds and rules can be fine-tuned through parameters in the code.

At this point a clear distinction for the boundary of separate characters for most of the given labelled data is possible. As such, the resulting data-label pairs are, with extremely high confidence, correct and contain one specific label per time series.\\
The remaining preprocessing pipeline is very straightforward and similar to one described in \citep{bulling2014tutorial} with normalization, resampling and, depending on the use-case, windowing of the data. Data is normalized column-wise to values between 0-1. Resampling is important, since the sensor values are recorded with varying frequency. The resample rate is chosen to be around 10 ms to capture small movements very well but to also avoid too much interpolation. Depending on the use-case of the data, a sliding window algorithm is also applied. The first use is simply as an input for the final classifier which can then take unlabelled time series data as an input and output a label. For this approach, sliding windows are taken with an overlap of about 25\%. To capture the process of writing a single symbol, a window size of around 16 is preferred, given the previous sample rate. The second use for the data is as an input for a different classifier with the aim to provide a way to perform label splitting on unknown sensor data for which the rule-based approach does not work since the number of labels has to be known. This use case does not require windowing as the amount of available samples is plentiful and the intent is not to classify regions but single samples.

\subsection*{Boundary data feature extraction and classification}
Considering the terms where the label splitting was successful we could then assign to each sample in such a series a value for if the person was in the process of writing a character. The idea is to use this information to somehow classify unknown samples into active or inactive, referring to the process of writing as being active. With a Deep Learning network already refined from tests for the final task, that is classifying the characters themselves, we tried using that network for this task as well. This would fail, however, mostly due to the network being extremely sensitive to the value of the force sensor. This outcome was not unlikely, since the data and labels were a direct result of the rule-based approach which used the force as a means of providing the labels we now use. As an alternative for a neural network we tried using Random Forest with various parameters which would provide results with similar issues. Concluding the tests, we went one step back and worked on feature engineering instead of using raw features, since we found that we would definitely need the force to produce meaningful results but could not find a way to scale or weigh it properly as a feature. The resulting idea was to use our, at this point during experimentation, already refined Deep Learning network, but instead of a classifier as a feature extractor only. The network is trained on the premise of trying to classify the active areas again, but during testing, instead of using the output of the final Dense layer, the output of the final LSTM layer is taken where each node is then treated as a feature of the input sample. The full designs of the network architectures are given in figure \ref{architectures} and get more in-depth explanations in chapter \ref{results}.\\
Given the engineered features for each sample, the Random Forest model is trained again, providing more accurate output this time. This model, called the boundary classifier, is then able to classify unknown samples of data as active or inactive after extracting their features. Inaccuracies during training can be cleaned up by applying similar rules based on domain knowledge that were used during the label splitting algorithm. As a result, a series of activity is then considered as one written character and can therefore be classified, without being certain about the amount of written symbols per term, or, in other words, the length of the label.

\subsection*{Segment and final term classification}
To finally classify the extracted regions, a model is trained by the formerly achieved properly segmented labelled data. Being as certain as possible about regions and their labels for most of the data, the network can be trained properly. Given that the boundary classifier and subsequent preprocessing produce region output similar to the training input, it is then possible to use this classifier to also classify completely unknown time series sensor data. Since every person's writing style is different the challenge permits five labelled terms of an unknown writer to be used to update the model before classifying unlabelled data from that person. The boundary data for the unknown writer is calculated through the boundary data classifier, as five labelled terms are not enough to apply the rule-based label splitting, and the resulting terms are then used to update the model for this writer specifically.

\section{Results and Evaluation}\label{results}
\begin{figure}[ht]
    \caption{Left: Preprocessing settings for training and testing data. Right: Real character to one-hot value mapping for reference.}
    \centering
    \includegraphics[width=\linewidth]{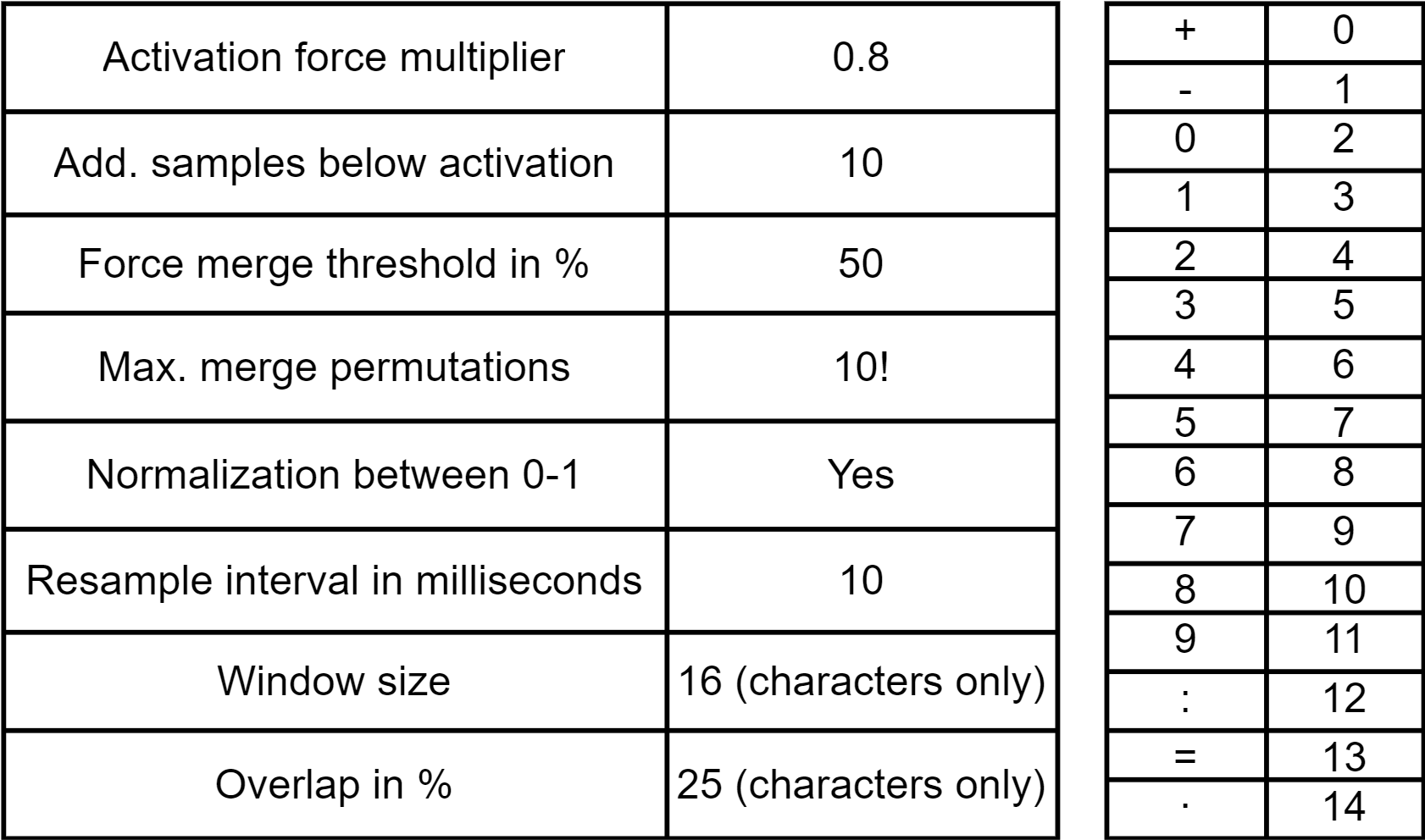}
    \label{preprocessing}
\end{figure}

In this chapter we outline the results for the training and testing of the boundary feature extractor, boundary classifier, character classifier as well as a simulation for the challenge. All networks are trained using data of all but one person that is picked randomly and is used for the simulation and therefore treated like an unknown writer. The trained networks require one-hot encoded class label input. Therefore the original classes are mapped to values. This mapping and the preprocessing settings are listed in figure \ref{preprocessing}.

The boundary feature extractor and character classifier models are trained with similar parameters albeit a slightly different architecture. Refer to figure \ref{architectures} for the exact designs. The architectures themselves are further explained in their respective sections. For the training process, the batch size is set to 2048 for the boundary feature extractor and 128 for the character classifier. Training is stopped if the validation loss does not decrease for more than five epochs. The weights of the epoch with the lowest validation loss are picked after stopping. The optimizer used is Adam with a learning rate of 0.001. Weights are regularized with l2 and a rate of 0.01. For the character classifier the convolution kernel size is set to [4, 1] in order to capture column-wise dependencies but ignore row-wise dependencies. Respectively the convolution kernel for Max Pooling is set to [2, 1] to cut the number of samples in half. For the boundary feature extractor the kernel size is always [1, 1] since the input is a single sample only. Finally, before connecting the last LSTM layer to the Dense output layer, a Dropout layer of 50\% is added. The Dense and Dropout layers are removed during testing of the boundary feature extraction.

\subsection*{Boundary classifier}
\begin{figure}[ht]
    \caption{These diagrams show the network architectures with input shapes and number of layers for the neural networks used in this work.}
    \begin{subfigure}[b]{0.23\textwidth}
        \includegraphics[width=\textwidth]{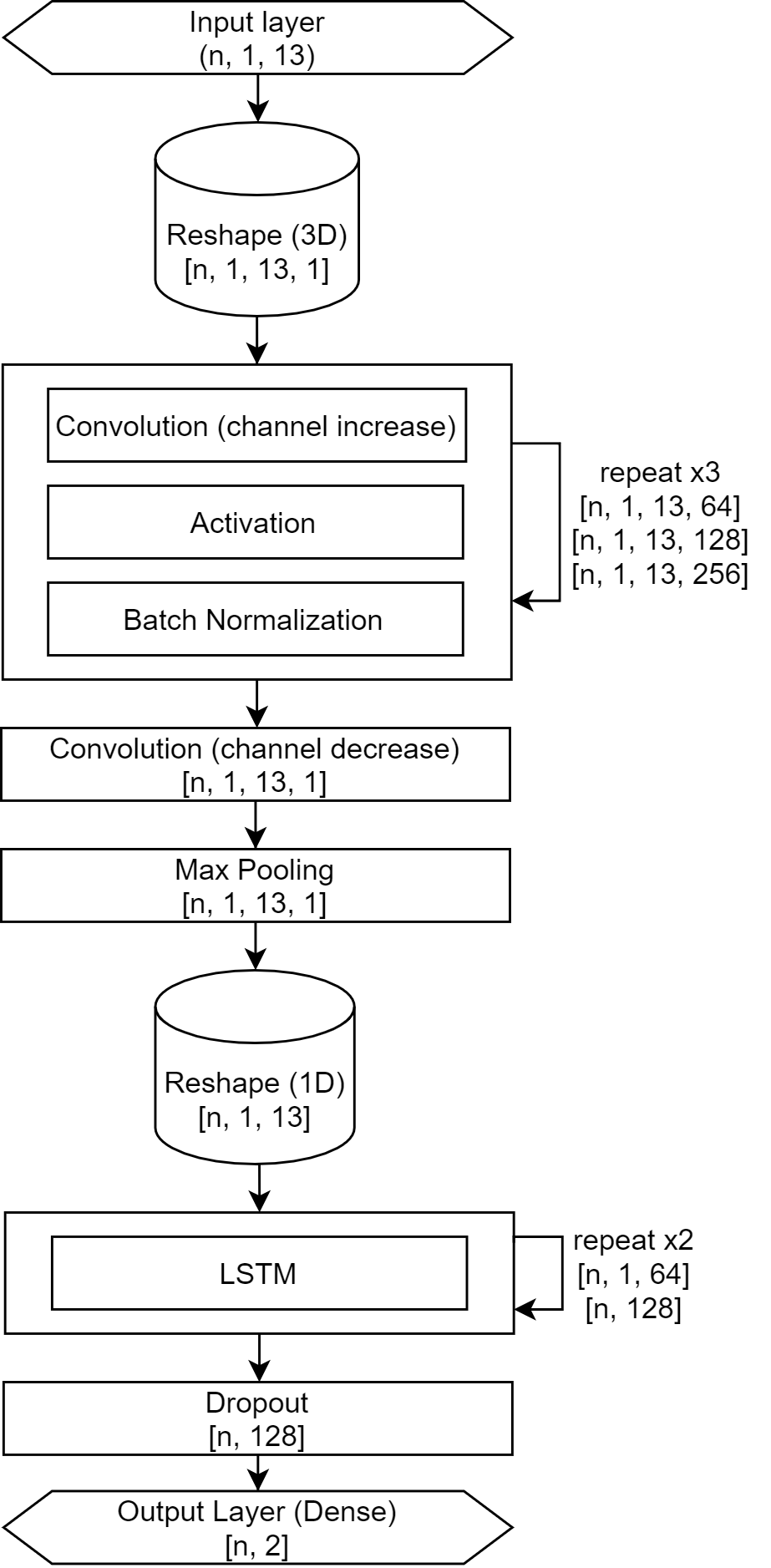}
        \caption{Network architecture for the boundary feature extractor. The kernel size for the convolution layers is set to [1, 1]. To extract features using this network, the last two layers are cut off during testing. The network learns to classify if a given sample is active (writing) or inactive (not writing).}
        \label{architecture_boundary_extractor}
    \end{subfigure}
    \hfill
    \begin{subfigure}[b]{0.23\textwidth}
        \includegraphics[width=\textwidth]{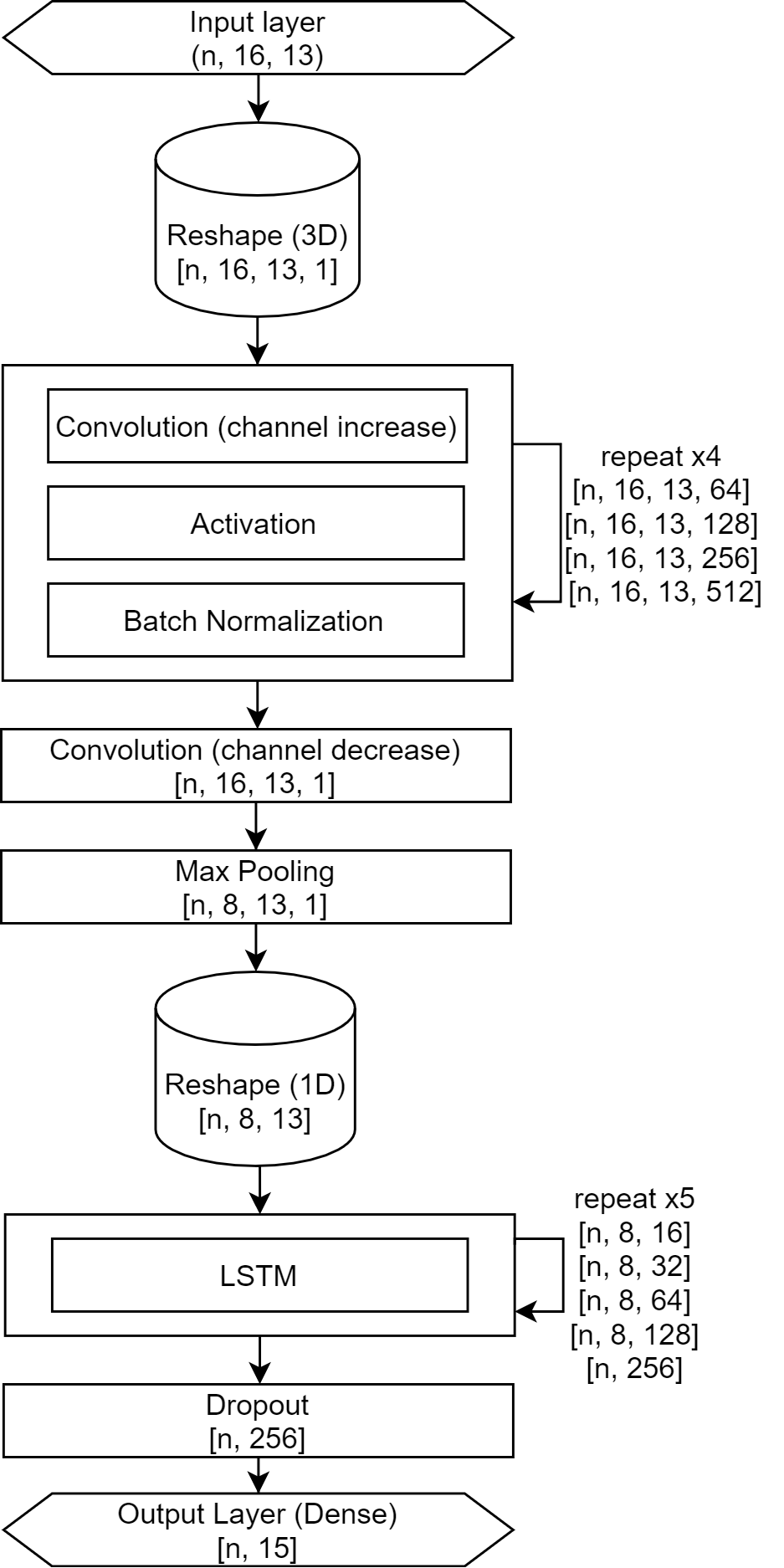}
        \caption{Network architecture for the character classifier. The kernel size for the convolution layers and Max Pooling layer is set to [4, 1] and [2, 1] respectively. The network learns to classify written characters from input windows. It is trained using data extracted through the label splitting algorithm.}
        \label{architecture_character_classifier}
    \end{subfigure}
    \label{architectures}
\end{figure}

The boundary data classifier is trained without windowed data. Instead singular samples and their label (active or inactive) are fed into the network. The data set is distributed in the following splits: \textbf{Total number of samples:} 4,500,977, \textbf{Train split:} 60\% (2,700,585), \textbf{Validation split and test split:} 20\% each (900,196 each). The exact network architecture itself is shown in figure \ref{architecture_boundary_extractor}.

The model loss stabilizes relatively quickly and the F1 score on the test set averages at slightly above 91\%. To extract features from boundary data it is then fed as test data into the network. The outputs of the final LSTM layer are taken as the extracted features and used as input for the Random Forest model. As a result of the network architecture there are 128 features per sample (previously 13). In the case of training, the extracted features are used with their original boundary labels. The number of estimators is set to 50 and only 25\% of the extracted features is randomly chosen for training. The number of active versus inactive samples is about equal. The resulting boundary classifier model also evaluates with a similar F1 score to the boundary feature extractor. However, the combination of both networks has proven to result in smoother predictions. This means that the amount of outliers is far fewer compared to only using one of the methods alone. This in turn makes the next step easier and cleaner, which is detecting and removing the remaining outliers in question to then generate windows and finally classify them.

\subsection*{Character classifier}
\begin{figure}[ht]
    \caption{These graphs show the performance of training and testing for the character classifier with the architecture shown in \ref{architecture_character_classifier}. Specifically the model loss and confusion matrix.}
    \begin{subfigure}[b]{0.31\textwidth}
        \includegraphics[width=\textwidth]{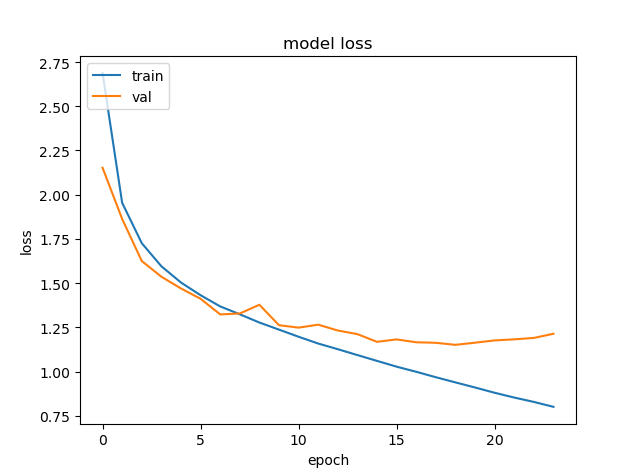}
        \caption{Model loss during training of character classifier. Both the training and validation loss decrease over time until the validation loss stabilizes after around 20 epochs when training is stopped to avoid overfitting.}
        \label{model_loss}
    \end{subfigure}
    \hfill
    \begin{subfigure}[b]{0.31\textwidth}
        \includegraphics[width=\textwidth]{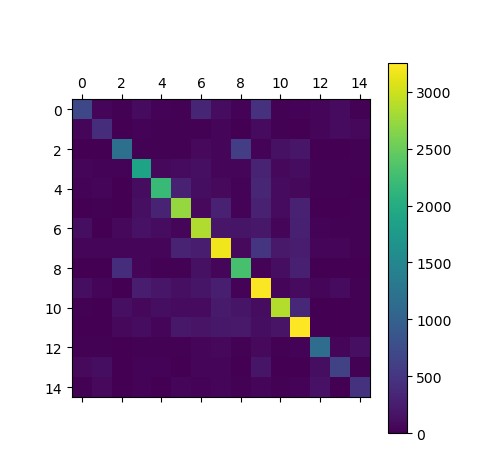}
        \caption{Confusion matrix using separate test set on the trained model. The diagonal is well visible. For classes 0, 1, 12, 13 and 14 fewer samples were available and tested, explaining the lower count in the matrix.}
        \label{confusion_matrix}
    \end{subfigure}
\end{figure}

To train the character classifier, windows with labels are required. From the segments acquired through label splitting, there are usually between 0-2 windows per segment. Given the previously stated preprocessing values, a segment needs to contain at least 16 samples to produce a window and at least 28 to produce another given the overlap, and so on. In rarer cases segments are too short, so they are discarded. This provides yet another opportunity to clean input data. The data set is distributed in the following splits: \textbf{Total number of windows:} 235,124, \textbf{Train split:} 60\% (141,074), \textbf{Validation split and test split:} 20\% each (47,025 each). During training the validation loss usually stopped decreasing after around 15-20 epochs for the character classifier. The exact network architecture itself is shown in figure \ref{architecture_character_classifier}. A graph depicting the model loss is shown in figure \ref{model_loss}.\\
Using the previously split test set, the F1 score of the model is evaluated. During five-fold cross-validation the F1 score ranges from 60\%-65\%. A confusion matrix for the test set is shown in figure \ref{confusion_matrix}. Refer to figure \ref{preprocessing} for the class to value mapping.

\subsection*{Challenge simulation}
Testing with completely unknown writers requires a few differences during preprocessing after boundary classification. For the challenge, five adaption samples for any unknown writer are given. These are regular labelled terms. We use these to adapt the character classifier with the intent for it to learn the writing style of the new person. Both the adaption and testing set undergo similar preprocessing steps than the training set, especially using the same parameters. Their boundary features are extracted and classified using the previously trained models. It is impossible to use the label splitting rule-based algorithm to determine the style of a writer from five terms. Therefore, these terms are also segmented using the boundary classifier. However, since the label is known, some rules can be applied. For both sets the boundary data is further cleaned. Outliers are generally removed by identifying singular samples or small groups of samples in between samples of the opposite group. Longer active regions are then extended by manually appending active or inactive labels to them, based on the distance to any other close active areas to avoid accidental segment merging. This extension of active regions is performed as to not miss truly active areas that were correctly recognized by the classifier but are too short to produce a window for character classification later on. We obviously saw an increase of classified characters through this extension, but also an increase in the score. The adaption set is further treated because the number of characters in the label is known and we can use this information to our advantage. After segmenting the data by the boundary information it is not guaranteed that the number of segments equals the number of actual characters. If we know the number of characters we can therefore merge segments to decrease the number of segments to fit the number of characters in the label. This merging is based on the distance between active areas. Finally, windows from the adaption and testing data can be generated. The adaption windows along with their true labels are then fed into the character classification model to update and adapt it to the new writer. The testing data is then evaluated on the updated model. The challenge itself is evaluated by the metric of Levensthein Distance which measures the distance between two input strings. With the randomly picked test person a total of 90 windows for adaption are created from the five adaption terms and the character classifier is updated on these. Notably, there are not samples for all available labels, so the model does not learn about the writing style for all characters of the person. A total of 236 terms is then evaluated on the character classifier with a resulting Levensthein Distance of ~8 and thus concluding the experiment.

\section{Conclusion}
In this work, state-of-the-art neural network designs are used to solve the common problem of classifying written characters. The data is captured by a regular shaped pen that is equipped with various sensors: the Stabilo DigiPen. Precisely, the final network was designed to be able to classify Arabic single digit integers and some of the most common mathematical operators. In addition to regular classification the task presented another challenge since the given sensor data streams are labelled but do not inherently provide boundaries between separate labels in the stream. The segmentation of characters in this stream on labelled data through our rule-based algorithm is fairly successful in identifying the writing style of a person and extracting correctly labelled segments through those rules. The yield averaged 70\% in this part of the workflow. The model trained on this data is able to classify characters independent of the writer with an F1 score of higher than 60\% at 15 classes. Because the segmentation of characters through the rule-based algorithm can not be performed on unlabelled data, another classifier is trained on the previously extracted data with the intent to classify the boundaries between characters. The resulting model classifies boundaries at an F1 score of higher than 90\% and also returns only few outliers that are cleaned easily. The segmentation of unlabelled data is performed with the boundary classifier and the segments are then predicted through the character classifier. During the challenge evaluation an adaption set is given for each writer so the character classifier can be updated using these terms beforehand. A simulation resulted in a Levensthein Distance of ~8 on a randomly picked test set.\\
Although the given data is fairly clean of other problems, the segmentation of characters is very challenging. Every person has a slightly different writing style and without any type of calibration or other additional information about the writers it is up to an algorithm to detect that style. We designed a straightforward rule-based algorithm which acts mainly on domain knowledge. Given the fairly high score of both trained classifiers we believe that we at least partially succeeded with that algorithm. On an unknown writer, the networks and some additional cleanup through topical knowledge and assumptions have to be enough to classify their written terms. Given the resulting Levenshtein Distance of our test it is quite clear that there are more required improvements to make this a feasible approach in a real-world scenario but our results show that the training works fairly well and may be used as a starting point for proper classification.\\
Given that the largest issue for this work are the aforementioned properties of the data-label pairs, we would like to suggest a simple improvement for the data capture: calibration. Letting the person write each symbol at least once, or better multiple times, as a calibration for the pen would only be required once but provides valuable knowledge for the classification task. Knowing the writing style of a person makes segmentation much easier and more accurate. Therefore, as we have shown with the label splitting algorithm, if the writing style is known, segmentation would be highly accurate and, as we have shown as well, classification could provide more accurate prediction as a result. We believe that this improvement would greatly enhance the results with our approach but, of course, other improvements like refining the networks or preprocessing algorithms may also lead to much better results without the requirement of additional data.
 
\bibliographystyle{ACM-Reference-Format}
\bibliography{literature}

\end{document}